\documentclass[letterpaper, 10 pt, conference]{ieeeconf}  

\IEEEoverridecommandlockouts                              

\usepackage{xcolor,soul,framed} 

\colorlet{shadecolor}{yellow}
\usepackage[pdftex]{graphicx}
\graphicspath{{../pdf/}{../jpeg/}}
\DeclareGraphicsExtensions{.pdf,.jpeg,.png}

\usepackage[cmex10]{amsmath}
\usepackage{array}
\usepackage{mdwmath}
\usepackage{mdwtab}
\usepackage{eqparbox}
\usepackage{url}
\usepackage{multirow}
\usepackage[thinlines]{easytable}
\usepackage{comment}

\title{\LARGE \bf
Novel Supernumerary Robotic Limb based on Variable Stiffness Actuators for Hemiplegic Patients Assistance}

\author{Basma B. Hasanen $^{1,2}$,
Mohammad I. Awad $^{1,2}$,
Mohamed N. Boushaki $^1$,
Zhenwei Niu $^{1,2}$,\\
Mohammed A. Ramadan $^{1,2}$,
and Irfan Hussain $^{1,2}$
\thanks{ $^1$Center of Autonomous Robotics Systems, Khalifa University,  Abu Dhabi, United Arab Emirates, P O Box 127788, Abu Dhabi, UAE}
\thanks{$^2$Mechanical Engineering Department, Khalifa University,  Abu Dhabi, United Arab Emirates, P O Box 127788, Abu Dhabi, UAE}%
\thanks{ Corresponding Author: Irfan Hussain (irfan.hussain@ku.ac.ae)}}

\providecommand{\keywords}[1]{\textbf{\textit{Index terms---}} #1}

\begin{document}
\maketitle
\thispagestyle{empty}
\pagestyle{empty}




\begin{abstract}

Loss of upper extremity motor control and function is an unremitting symptom in post-stroke patients. This would impose hardships on accomplishing their daily life activities. Supernumerary robotic limbs (SRLs) were introduced as a solution to regain the lost Degrees of Freedom (DoFs) by introducing an independent new limb. The actuation systems in SRL can be categorized into rigid and soft actuators. Soft actuators have proven advantageous over their rigid counterparts through intrinsic safety, cost, and energy efficiency. However, they suffer from low stiffness, which jeopardizes their accuracy. Variable Stiffness Actuators (VSAs) are newly developed technologies that have been proven to ensure accuracy and safety. In this paper, we introduce the novel Supernumerary Robotic Limb based on Variable Stiffness Actuators. Based on our knowledge, the proposed proof-of-concept SRL is the first that utilizes Variable Stiffness Actuators. The developed SRL would assist post-stroke patients in bi-manual tasks, e.g., eating with a fork and knife. The modeling, design, and realization of the system are illustrated. The proposed SRL was evaluated and verified for its accuracy via predefined trajectories. The safety was verified by utilizing the momentum observer for collision detection, and several post-collision reaction strategies were evaluated through the Soft Tissue Injury Test. The assistance process is qualitatively verified through standard user-satisfaction questionnaire.

\end{abstract}

\vspace{0.3cm}



%
\IEEEpeerreviewmaketitle

\section{Introduction}

Nowadays, the world witnesses high rates of strokes affecting youth and the elderly. Based on the Global Burden of Disease and Risk Factors Study report, there have been 12 million different stroke cases worldwide in 2019 \cite{johnson2019global}. Strokes result in serious health problems ranging from temporary partial to lasting total disability of one or more of a human's limbs. One of the possible post-stroke impairments is the total functional disability of one of the upper limbs. Based on  \cite{kwakkel1999intensity}, restoring the lost upper limb functions is more difficult than restoring lower limb functions in post-stroke patients. Besides, it was reported that only 14\% of post-stroke patients recover from a stroke back to their original healthy state, 25\% have a level of enhancement in their affected extremities, and nearly 60\% suffer from lasting complete limb malfunction \cite{wade1983hemiplegic}. As a result, hemiplegic patients spend more time completing regular daily life tasks. Besides, normally bimanual performed tasks would cause fatigue to the healthy limb, negatively impacting the patient's physical and psychological states \cite{8324526}. 

\begin{figure} 
\centering
\includegraphics[width=2.4in]{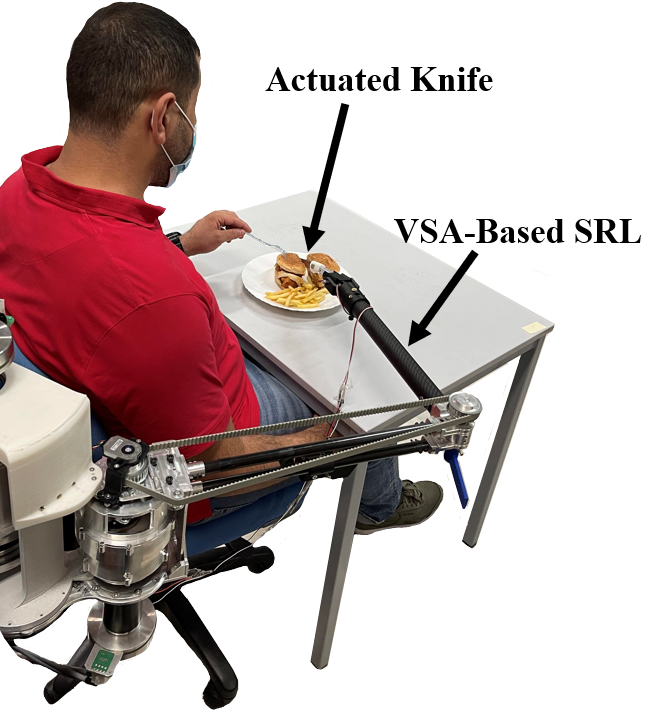}
\hfil
\caption{The proposed  SRL  with variable stiffness joints to improve both the hemiplegic patient's safety and the bimanual eating task efficiency. The system is fixed to the chair.}
\label{fig_rend}
\vspace{-0.6cm}
\end{figure}

Recently, Supernumerary Robotic Limbs (SRLs) have been introduced as a solution to assist hemiplegic patients in overcoming the difficulties associated with activities of daily life (ADL) \cite{salvietti2016compensating,MITSRF_ort2015supernumerary}. This assistance can be in the form of augmenting the healthy limb (arm or finger) by expanding its capabilities towards completing bimanual tasks or compensating for the missing manipulation capabilities of the impaired limb by adding an extra limb \cite{hussain2015using}. The significance of SRLs over other assistive robotic solutions (i.e., exoskeletons and prostheses) lies in their independence from the presence (or absence) of the impaired limb (unlike prostheses) and their independence from being constrained by the limbs' kinematics (unlike exoskeletons) \cite{hussain2017toward}. In the literature, supernumerary robotic fingers were proposed for augmenting the healthy arm of hemiplegic patients to enable them to perform bimanual tasks \cite{MITSRF_wu2014supernumerary,MITSRF_wu2015hold,hussain2020enhancing}. They were also proposed to compensate for the missing manipulation capabilities in the impaired hand \cite{hussain2016soft,hussain2017soft}. MIT recently proposed supernumerary robotic arms for hemiplegic patients to compensate for the impaired arm's capabilities in the bimanual task of eating with a fork and a knife \cite{9353202}. In their work, they proposed a control strategy that integrated voluntary human movements and the autonomous reactive control of a robotic limb. Apart from the interfaces, another challenge lies in the design for safe interaction with humans during task execution without jeopardizing the accuracy or the task completion time. In our work, we are proposing the utilization of Variable Stiffness Actuators (VSAs) in supernumerary robotic arms to harvest the advantageous features of soft actuators in their intrinsic safety and energy efficiency, and the advantageous features of rigid actuators in their capability to perform high-accuracy tasks through the capability of VSA in tuning the stiffness to high levels \cite{vanderborght2013variable}. In the literature, a broad spectrum of variable stiffness actuators was proposed \cite{19_VSA_TONIETTI2005,20_VSA_SCHIAVI2008,22_VSA_jafari2010,35_VSA_Sun2017}, the motivation behind each proposed design may include the stiffness range, the rate-of-change in stiffness, the maximum elastic deflection, the maximum elastic-energy stored, and energy-efficiency. To the author's knowledge, a supernumerary robotic limb with variable stiffness actuators has never been proposed.

In this paper, we propose the first supernumerary robotic limb based on VSAs (see Fig. 1). The proposed system is a proof-of-concept for a two-Degrees-of-Freedom (two-DoFs) SRL that would be utilized for the assistance of post-stroke patients in bimanual tasks (i.e., eating with a fork and knife). The contributions of this manuscript lie in: (1) the design and development of the VSA-based-SRL, including the modeling, analysis, mechanical design, and realization. (2) the evaluation process which aims to validate the system's accuracy and safety. The system's accuracy was evaluated by subjecting the SRL to predefined trajectories and measuring how accurately it followed these trajectories. In order to ensure the safety of the system in case of accidents (i.e., collisions), the momentum observer was selected for collision detection, and multiple post-collision reaction strategies were proposed and evaluated based on the Soft-Tissue Injury Test \cite{haddadin2010soft}. Afterward, a state-machine control system was developed to perform the process of assistance in eating with a fork and knife. Finally, a satisfaction questionnaire was conducted with healthy users to provide remarks for further development prior to moving toward hemiplegic patients.

The rest of the paper is organized as follows: Section \ref{sec2:SRL Modeling} illustrates the workspace analysis and dynamic modeling. The system realization and actuation system are illustrated in Section \ref{sec3:DesignandDevelopment}. The experimental setup and the evaluation experiments for accuracy and safety are illustrated in Section \ref{sec4:ExpAndEva}. The qualitative experiment is depicted in Section \ref{sec5:qualititative}. Finally, the conclusion and future work are illustrated in Section VI.

\vspace{-0.1cm}
\section{Supernumerary Limb Modeling}
\label{sec2:SRL Modeling}
The dynamic modeling and the workspace analysis are illustrated in this section. The dynamic model will be utilized in the safety evaluation experiment as the selected collision detection method is model-based. At the same time, the workspace analysis would determine several design parameters, such as the degrees of freedom, the lengths of the links, and joint limits.

\begin{figure}
\centering
\includegraphics[width=3.15in]{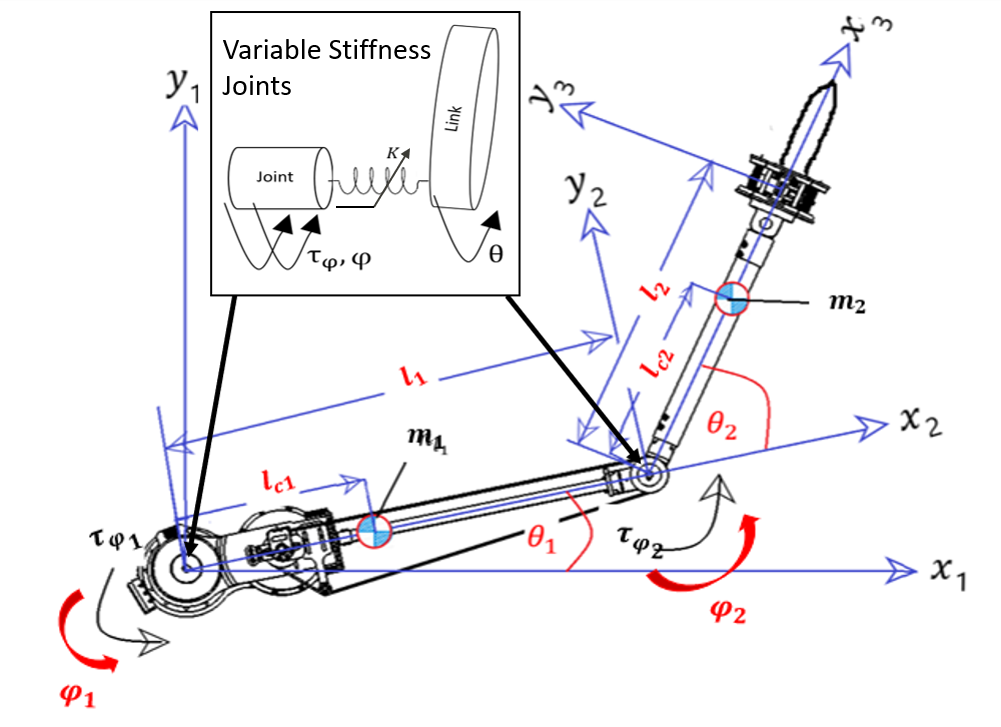}
\caption{Schematic of the proposed VSA-based-SRL for Dynamic Modeling.}
\label{fig_sketch}
\vspace{-0.6cm}
\end{figure}

\subsection{Dynamics Model}
\label{sec:dynamic model}
A simplified schematic of the SRL is shown in Fig. \ref{fig_sketch} and used to derive the SRL's equations of motion (EOMs). The system consists of two links with variable stiffness joints. A simplified dynamic model is obtained by considering the VSA system as a two-inertia-spring-damper system where the spring's stiffness can be changed

\cite{vanderborght2013variable}. Hence, the system can be represented by four generalized coordinates: ($\theta_{1}$, $\theta_{2}$, $\phi_{1}$, $\phi_{2}$), which are the links and joint motors' rotational positions, respectively. All annotations are depicted in Fig. \ref{fig_sketch}. The Euler-Lagrange method was used to obtain the standard equations of motion as in \cite{spong2008robot}:

\begin{equation}
\label{eq:dynamic equation}
\begin{aligned}
M_{1}(\theta)\ddot{\theta} + C({\theta}, {\Dot{\theta}}) + g(\theta) + K (\theta-\phi) + F_{rl} &= \tau_{ext}\\
M_{2}\ddot{\phi} + K (\phi-\theta) + F_{rm} &= \tau_{m}
\end{aligned}
\end{equation}


Where, $M_{1}(\theta)$, $M_{2}$ are inertia matrices for link and motor sides respectively, $C(\theta,\Dot{\theta})$ signifies the Coriolis and centrifugal torque matrix, $g(\theta)$ represents the gravitational torque terms, and $K$ is the stiffness matrix. $F_{rl}, F_{rj}$ are the frictional vectors opposing the motion of both the links and joints, respectively. $\tau_{ext}$ is the external torque, and $\tau_{m}$ is the motor torque.

\subsection{Workspace Analysis and Kinematics Modelling}
\label{sec:workspace}
\label{sec:workspace}

The dimension of the SRL workspace is one of the main criteria to be considered during the robot design process to meet the task's functional requirements \cite{siciliano2016springer}. The bimanual eating task using a knife and a fork can be sufficiently accomplished by the hemiplegic patient using his non-paralyzed arm, and a two-DoFs robotic arm \cite{9353202}. 

To achieve the task, the patient holds the fork, and the knife is attached to the robot end-effector. The two-DoFs robotic limb positions the end-effector at the desired position close to the plate while the actuated knife is utilized for cutting. The operational plane of the SRL arm is tilted at a slight angle from the horizontal plane.

\begin{figure} [t!]
\centering
\includegraphics[width=3.5in]{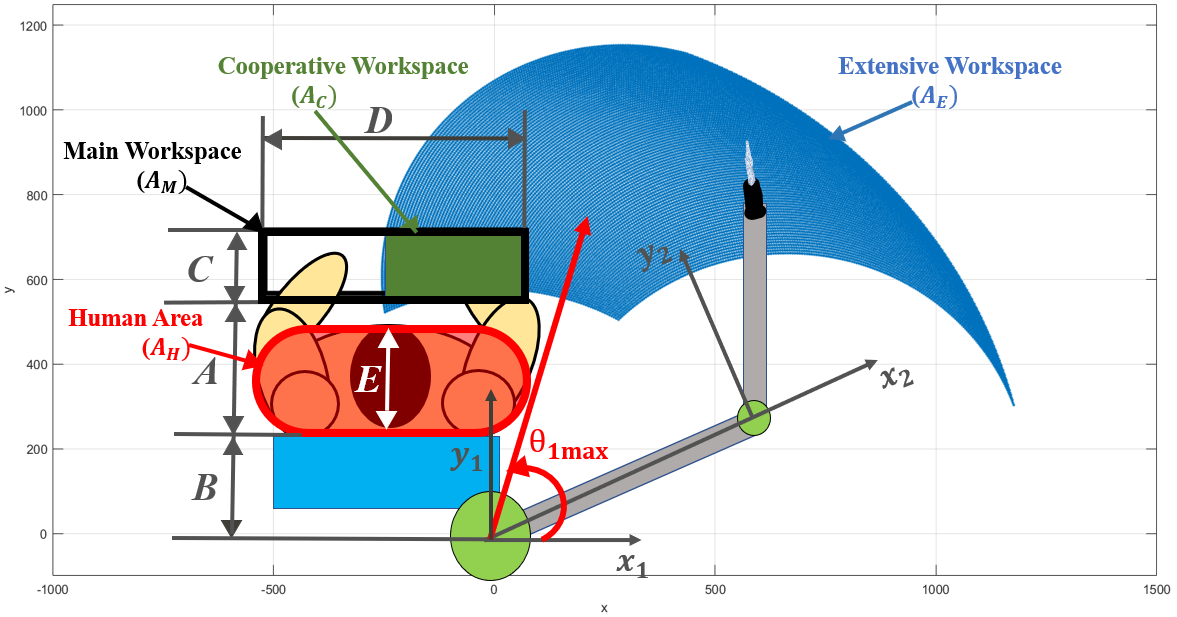}
\hfil
\vspace{-0.5cm}
\caption{The top view of the SRL's optimized workspace using MATLAB, the dimensions A, B, C, D, and E are based on the users' anthropometric dimensions, the whole body kinematics during eating, and the dimensions of the harness and the actuators. The different colors can be interpreted as follows:
(1) The 'black' rectangle shows the main workspace of the user (required for the bimanual eating task).
(2) The 'green' area represents the cooperative workspace.
(3) The 'blue' area is the extensive workspace of the SRL.
(4) The 'red' area represents the area occupied by the human body.}

\label{fig_work}
\vspace{-0.5cm}

\end{figure}

According to \cite{nakabayashi2017development}, the main human and SRL workspaces can be classified into extensive, cooperative, and invasive. Fig. \ref{fig_work} shows the cooperative and extensive workspaces in different colors. 
Referring to \cite{jacobs2008ergonomics}, ensuring the assistive device will accommodate sizes ranging from the $5^{th}$ percentile female to the $95^{th}$ percentile male of the targeted users is good practice. The anthropometric dimensional data from \cite{jacobs2008ergonomics} along with the recorded whole body kinematics during eating from \cite{nakatake2021exploring}  were utilized to estimate the following dimensions in Fig. \ref{fig_work}: 
  \begin{enumerate}
          \item $A$ is the minimum horizontal reach required by the $5^{th}$ percentile female, ($A$ $= 327$ $mm$). 
          \item $C$ is the range of horizontal reach from $5^{th}$ percentile female (minimum) to $95^{th}$ percentile male (maximum), ($C$ $= 150$ $mm$). 
          \item $D$ is the maximum workspace breadth necessary for both $5^{th}$ percentile female and $95^{th}$ percentile male to complete the eating task, ($D$ $= 589$ $mm$).
          \item $E$ is the $95^{th}$ percentile male chest-depth, ($E$ $= 280$ $mm$).
    \end{enumerate}

The dimension $B$ $= 240$ $mm$, depends on the size of both harness and VSAs. Thus, the extra robotic arm's minimum and maximum required reaches are $567$ $mm$ ($A$+$B$) and $717$ $mm$ ($A$+$B$+$C$), respectively, defined in the base frame.

The problem of optimizing the SRL's workspace ($A_{W}$) can be formulated as follows:
 
\begin{itemize}
    \item The decision variables are: (${l_{1}}$, ${l_{2}}$, $\theta_{1_{max}}$, $\theta_{2_{max}}$) which are links (1, 2) lengths, and Joints (1, 2) limits, respectively.
    
    \item The constraints are: \textbf{(1)} the link 1 of the SRL should not reach the human area $A_H$ (red area) during the operation, this imposes $\theta_{1_{max}} \leq 65$ degrees; \textbf{(2)} the cooperative workspace $A_C$ (green area), which is a subspace of the main workspace $A_M$ (the black rectangle), should be included in $A_W$: $A_C$ $\subset $ $A_W$. The remaining area from $A_M$ is reachable by the human hand, with minimum accessibility by the SRL for safety purpose. \textbf{(3)} The SRL end-effector should not reach the human area $A_H$ (red area) during the operation; this means: $A_W$ $\cap$ $A_H$ = $\emptyset$. 

    \item The objective of the optimization is to satisfy the previous constraints with the minimum lengths of both SRL links. 
\end{itemize}

As depicted in Fig. \ref{fig_sketch}, frames were assigned to each joint; then, the forward kinematics was implemented following the Denavit–Hartenberg convention. Finally, the brute force optimization method was used in which all the ranges of the decision variables were swept in search of optimal values. The optimization results are as follows: ${l_{1}}$ $ = 674$ $ mm$ and ${l_{2}}$ $ = 545$ $ mm$, $\theta_{1_{max}} = 65 ^{\circ}$ and $\theta_{2_{max}} = 125 ^{\circ}$.

\section{Design and Development}
\label{sec3:DesignandDevelopment}
\begin{figure} 
\centering

\includegraphics[width=3in]{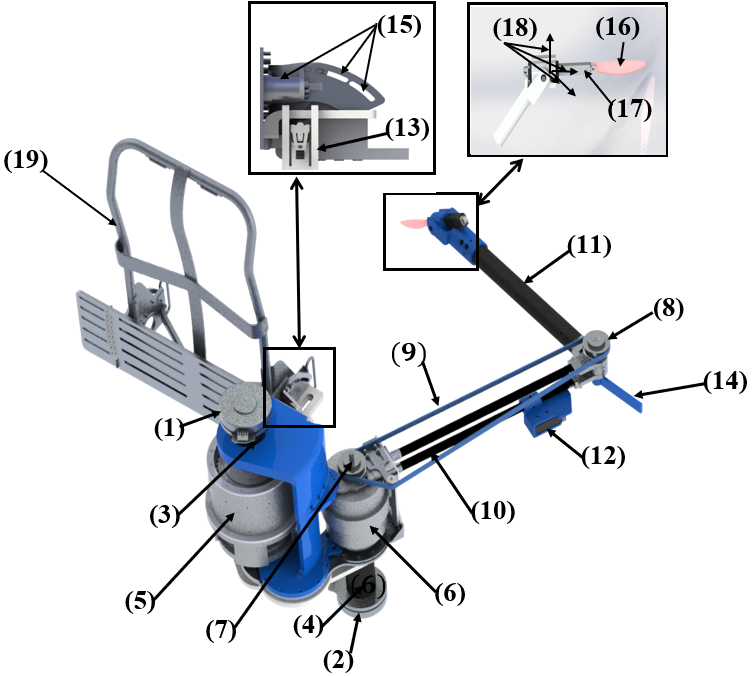}

\hfil

\caption{The CAD model of the SRL: (1), (2) EC 90 Motor.
(3), (4) Gearbox (91:1 gear ratio).
(5) VSA 1.
(6) VSA 2.
(7), (8) Pulleys (T5 Series Timing belt pulleys).
(9) Timing Belt (T5 Series).
(10) First Link (Carbon Fiber Tubes).
(11) Second Link (Carbon Fiber Tube).
(12), (13) Limit Switches.
(14) Limiter.
(15) Tilting Mechanism.
(16) Knife blade.
(17) ACTUONIX Linear Actuator L16.5 mm.
(18) Adjustable Axes.
(19) Harness.
}
\label{Fig:CAD_SRL_model}
\vspace{-0.6cm}
\end{figure}

The design and development of the VSA-Based-SRL have three main components: the development of the Variable Stiffness Actuators, the design of the two-DoFs arm, and the development of the end-effector. 

The design of the utilized Variable Stiffness Actuators was adopted from work proposed by our team in \cite{awad2016modeling}. The actuator tunes the stiffness by altering the transmission between the elastic element and the output link; this has been realized through a line mechanism that allows the tuning of stiffness to physically achieve a wide spectrum from low levels ($70$ $N.m/rad$) to significantly high levels ($8000$ $N.m/rad$) within minimal stiffness altering time ($450$ $ms$). When operating at low stiffness, safety can be ensured through the capability of shock absorption due to the actuator's elasticity. When operating at high stiffness levels, accuracy can be ensured as the utilized VSA operates similarly to a rigid actuator. Akin to the majority of the proposed VSAs, the utilized VSAs have two motors; the first motor (Maxon Brushless DC Motors EC90 with planetary gearbox 91:1 ratio, embedded 2048 Pulses-per-Revolution (PPR) resolution encoder, and EPOS Drive Controller) is responsible for altering the equilibrium point of the output link, while the second motor (Dynamixel M64 with Arbotix Controller) is responsible for driving the stiffness tuning line-mechanism. In addition, a linear potentiometer is utilized as feedback for the states of the line mechanism.

The mechanical design of the SRL arm is depicted in Fig. \ref{Fig:CAD_SRL_model}. The figure shows that the SRL consists of two links driven by Variable Stiffness Actuators (VSA 1 and VSA 2). The length of the links was defined as per the previous section. Moreover, to minimize the SRL's weight, the links were made from Carbon-Fiber tubes and Aluminum. Furthermore, to minimize the inertia of the SRL, the actuator's location was selected to be nearest to the base. The first actuator is mounted directly on the base, and its shaft holds the second actuator and the first link. The second actuator controls the second link through the timing-belt mechanism. The mount of the SRL was made of Aluminum. The mount has linear and tilted slots to allow the user to fix the SRL to the desired operational plane manually. Two limit switches were added to create a homing routine. The homing routine is necessary as the joint angles are measured via incremental encoders (CUI AMT102V with 2048 PPR resolution).

The end effector consists of an actuated knife mounted on a passive three-DoFs spherical joint whose axes intersect at a single point. The patient can adjust the knife's orientation using their healthy hand before cutting. The linear actuator is responsible for generating reciprocating movements of the knife, which the patient controls through a push-button pressed by their foot. The design specifications of the SRL are illustrated in Table \ref{VSADesignParameters}.

 \begin{table}[t!]
 \caption{The design parameters and main characteristics of the proposed VSA-Based-SRL}
 \label{VSADesignParameters}
 \centering
 \begin{tabular}{|p{0.2\linewidth} | p{0.2\linewidth}|p{0.2\linewidth} |p{0.15\linewidth}|}
 \hline
 Parameter & Value & Parameter & Value\\ 
 \hline
 Joint 1 Range of Motion & $0-65$ [$deg$] & Joint 2 Range of Motion  & $0-125$ [$deg$]\\
 \hline
 Joints Maximum Torque & $35$  [$Nm$] & Joints Maximum Speed & $120$ [$deg/s$]\\
 \hline
Joints  Range of Stiffness &  $70-8000$   [$Nm/rad$] &  Nominal Stiffness Variation Time & $450$ [$ms$]\\
 \hline
Link 1 Length & $674$ [$mm$] & Link 2 Length & $545$ [$mm$] \\
 \hline
 \end{tabular}
\vspace{-0.5cm}
 \end{table}

\vspace{-0.1cm}
\section{Experimental Work and Evaluation}
In this section, the experimental setup and the evaluations for safety and accuracy are illustrated. The experimental setup elucidates the integrated mechatronic system and the data exchanged among the system's components. Then, the safety evaluation experiment is conducted to define the maximum operating velocity such that the safety of the user is ensured during the compensation task. Lastly, the system's accuracy is evaluated through the trajectory tracking experiment.

\label{sec4:ExpAndEva}

\vspace{-0.1cm}
\subsection{Experimental Setup}
\begin{figure} 
\centering
\includegraphics[width=3in]{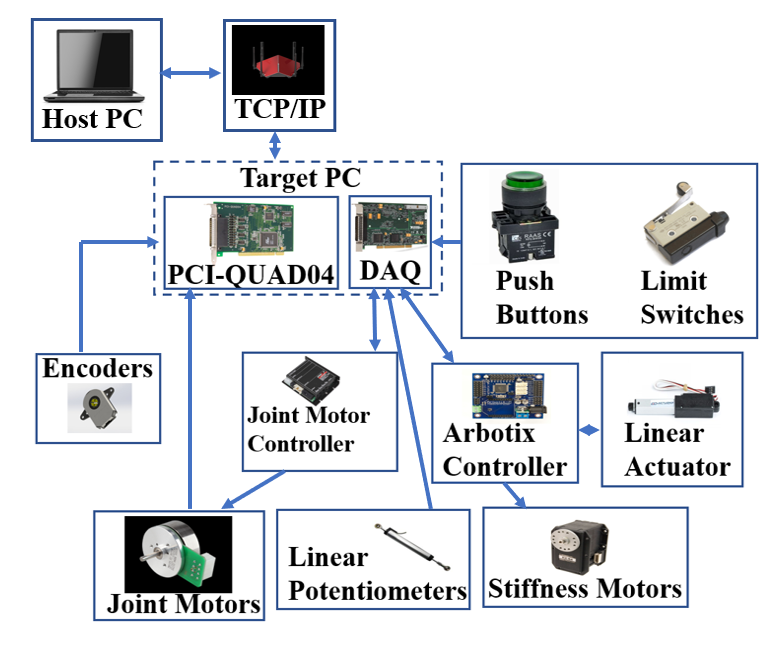}

\hfil

\caption{The experimental setup used to control the SRL. The connections between various components are demonstrated.}
\label{FigExp}
\vspace{-0.6cm}
\end{figure}

The experimental hardware setup incorporates several components, which are integrated as elucidated in Fig. \ref{FigExp}. It consists of the host computer and the target computer, which communicate through TCP/IP. The target computer contains the data acquisition device (DAQ: NI PCI-6221M) and the Quadrature Encoder Board (PCI-QUAD04). This DAQ communicates with the drivers of the VSA motors (joint motors (Maxon EC90) and stiffness tuning motors (Dynamixel M64) are driven by EPOS 24/5 and Arbotix, respectively). Moreover, the DAQ is also connected with limit switches, linear potentiometers, and push buttons. The Quadrature Encoder Board is connected to the embedded encoder in Maxon EC90 motors and the joint encoders. The linear actuator of the knife is controlled through the Arbotix driver. 

The controller program was implemented in MATLAB/SIMULINK environment on the host computer. The controller is transferred to the target computer and executed in a Simulink Real-Time environment (MATLAB xPC). The TCI/IP communication between the host and target computer operates at 100 Mbps. The DAQ executes the controller and communicates its command signals to the motor drivers through analog inputs (for the EPOS), PWM Digital Outputs (for the Dynamixel), and one digital output for the linear actuator, which is preprogrammed to perform reciprocating movements through the Arbotix. Meanwhile, the feedback data from the linear potentiometers are read by the DAQ through the Analog Inputs, while the limit switches and push buttons are read as Digital Inputs. The motor positions are read through the Quadrature Encoder Board. The data exchange rate between the sensory system, motor drives, and the target computer is 1000 samples per second.

\vspace{-0.1cm}

\subsection{SRL Safety Evaluation}
During the compensation process of the SRL, the patient's safety must be ensured, mainly due to the proximity of the SRL to the hemiplegic patient. Operating at low stiffness would ensure safety as it would enable the feature of shock absorption. Compared with rigid actuator-based SRL, the proposed SRL would be able to operate at higher velocities with a similar level of safety, which would enhance performance. In order to ensure safety in post-collision scenarios, the collision should be detected, and a post-collision reaction strategy should be followed. This work adopted a momentum observer for collision detection, and several reaction strategies were evaluated via the soft tissue injury test. The detailed illustration of the process is as follows:

\subsubsection{Collision Detection}
The momentum observer \cite{haddadin2017robot} is used for detecting collisions during Human-Robot Interaction (HRI) applications. Based on the dynamic model in Section \ref{sec:dynamic model}, the momentum observer can be designed as follows:

\begin{equation}
\begin{aligned}
    r=K_{O}[&p(t)-p(0)\\
    &-\int_{0}^{t} (\tau_{\phi}-F_{rl}-F_{rj}-\hat{\beta}(\theta, \dot{\theta})+r) dt],
\end{aligned}
\end{equation}
where $p$ is the momentum, which is defined as $p=M_{1}(\theta)\dot{\theta}+M_{2}\dot{\phi}$, and $\beta(\theta, \dot{\theta})=g({\theta})-C_{\theta}^{T}\dot{\theta}$. $K_{O}$ is a positive diagonal gain matrix.

The residual signal $r$ is used to detect the collision. In the experiments, selecting an appropriate threshold is important to increase collision detection accuracy. According to \cite{haddadin2017robot}, the threshold of the residual $r$ is set as follows:
\begin{equation}
\label{eq:threshold of r}
    \epsilon_{r}=\hat{r}_{max}+\epsilon_{c},
\end{equation}
where $\hat{r}_{max}$ is the maximum value of the collision detection signal during the movement of the SRL without any external collision event. $\epsilon_{c}$ is a small constant to avoid false detection.

As soon as the residual signal $r$ exceeds the threshold (${\epsilon}_{r}=8-20$ N.m), it activates the subroutine program that contains the post-collision reaction strategy which aims to minimize the harm to human users and the robot structures. Further details are illustrated in the following subsection.

\subsubsection{Post-Collision Reaction Strategies}
In order to manifest the advantages of utilizing variable stiffness actuators in SRL systems, two different post-collision reaction strategies are proposed. The zero-torque mode is the first reaction strategy where the SRL would be stopped by providing no input torque to the motors once the collision is detected. In this strategy, the level of stiffness is constant (high or low). The second strategy would combine the zero-torque mode with a rapid shift from high to low stiffness once the collision is detected.

\subsubsection{Soft-Tissue Injury Test}
\begin{figure} [t!]
\centering
\includegraphics[width=3.5in]{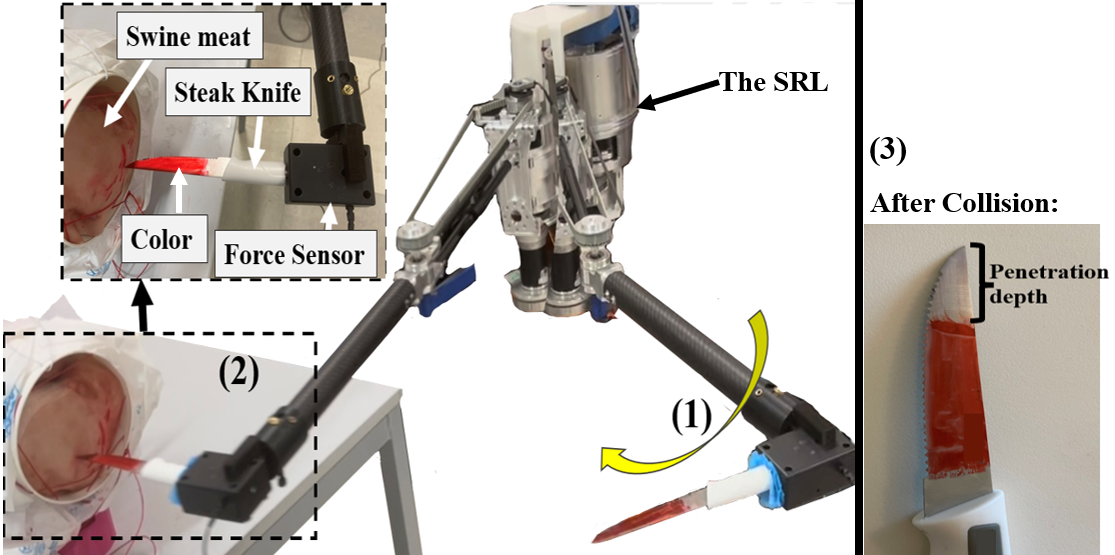}
\hfil
\caption{The safety experiment procedure with different components annotated: (1) The SRL would be set on a predefined trajectory to perform stabbing. The knife is rigidly attached to the Force sensor (Vernier dual-range), and the blade is colored with a special color (red). (2) A piece of swine meat is placed along the SRL's predefined trajectory such that it would be stabbed by the steak knife (11CM size). (3) After the collision, the penetration depth is measured manually and indicated by the wiped red color from the knife; the force value is read from the force sensor.}
\label{fig_safe}
\vspace{-0.5cm}
\end{figure}

The soft tissue injury tests inspired from \cite{haddadin2010soft} were explicitly chosen and implemented to assess the possible risk associated with using a sharp knife. The test aims to define the maximum operating velocity of the end-effector such that it would avoid any penetration of the knife into the patient's body. Through this test, the SRL would be equipped with a sharp tool (steak knife) and programmed to stab a piece of swine meat at different stabbing velocities and operating stiffness levels. The experimental setup is depicted in Fig. \ref{fig_safe}. The two criteria acquired in this test to assess safety are the penetration force ($ F_{p} $) and depth ($ d_{p} $). As shown in Fig. \ref{fig_safe}, a force sensor (Vernier dual-range force sensor) mounted on the end-effector is used to measure the penetration force. In addition, the knife's blade was painted with special paint that will be wiped away during the penetration to measure the penetration depth. 

In order to evaluate the safety of VSA-based-SRL and compare it with rigid-actuator-based SRLs, the SRL performed the stabbing of the swine meat at different velocities for three cases; (1) operating at low stiffness ($70 Nm/rad$)  $\rightarrow$ collision detected $\rightarrow$ zero-torque mode strategy activated. (2) Operating at high stiffness ($8000 Nm/rad$) $\rightarrow$ collision detected $\rightarrow$ zero-torque mode reaction strategy activated, and (3) operating at high stiffness $\rightarrow$ collision detected $\rightarrow$ (zero-torque mode and rapid decrease in stiffness level strategy activated). It is worth mentioning that case (2) is performed as a reference to compare the performance of the VSA with rigid actuators.  

\begin{figure} 
\centering
\includegraphics[width=2.9in]{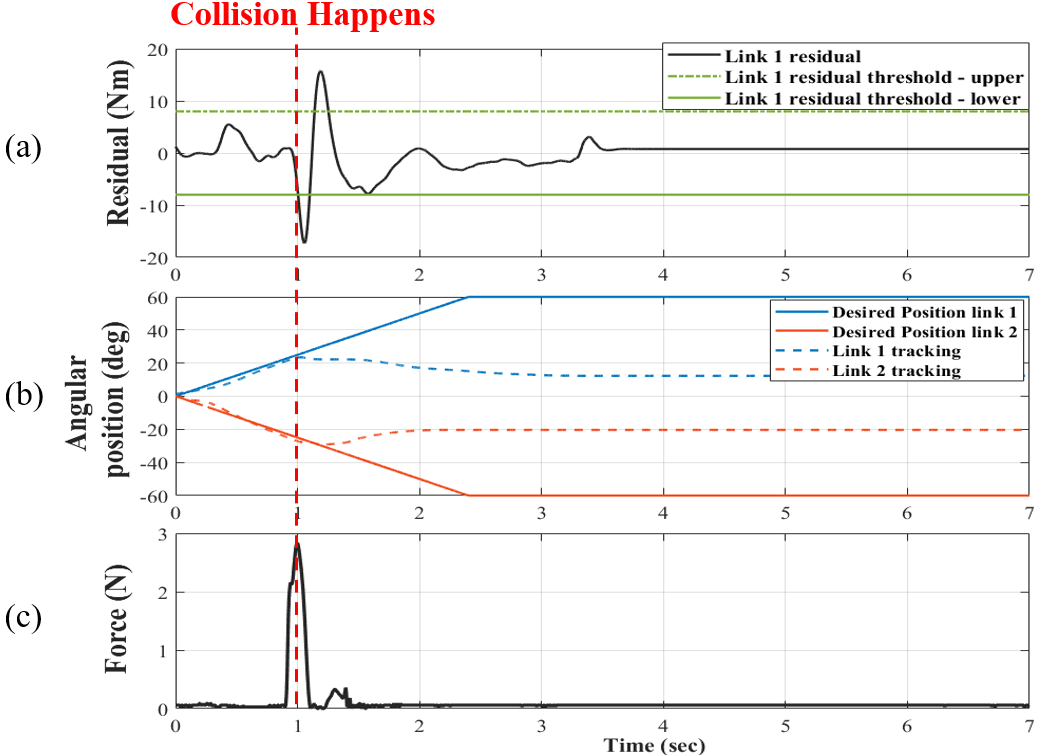}
\hfil

\caption{Results for forced dynamic collision (stabbing experiment). (a) The momentum observer residual signal to detect the collision, (b) the desired and actual trajectory of the SRL, (c) the Penetration Force. }
\label{fig_residual_results}
\vspace{-0.2cm}
\end{figure}

Figure \ref{fig_residual_results} illustrates the results of the forced dynamic collision at low stiffness mode (case (1)). The momentum observer's residual signal ($r$) is under the threshold before the collision occurs, and the manipulator keeps tracking the reference (see Fig. \ref{fig_residual_results}a). Once the collision is detected, the system is switched to the zero-torque mode, which would stop the manipulator and prevent it from following its desired trajectory (see Fig. \ref{fig_residual_results}b). The penetration force is recorded during the collision (see Fig. \ref{fig_residual_results}c), while the penetration depth is measured manually after the collision.

\begin{figure} 
\centering
\includegraphics[width=3in]{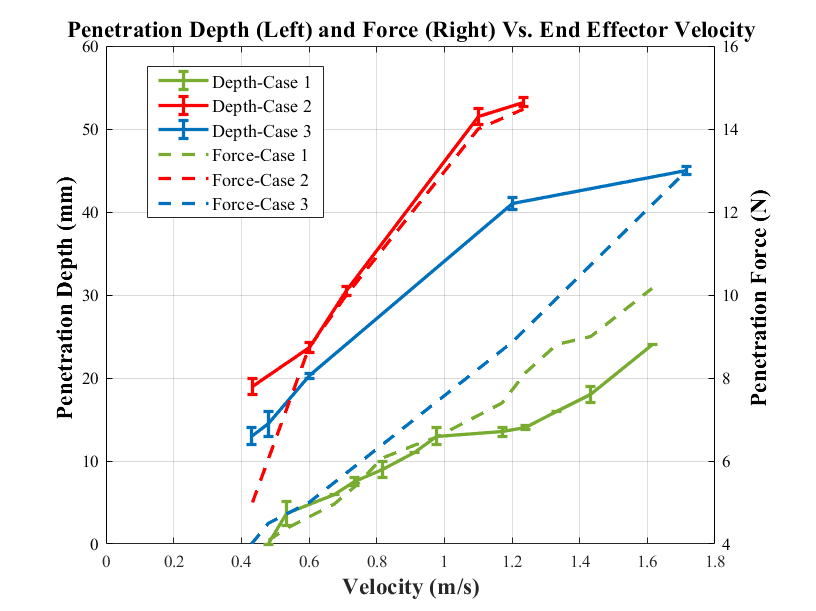}
\hfil
\caption{Penetration Depth (Left y-axis), and Penetration Force (Right y-axis) are plotted versus the end-effector velocity}
\label{fig_PenFor}
\vspace{-0.6cm}
\end{figure}

Similar experiments were conducted for the three cases at different velocities. The results are illustrated in Fig. \ref{fig_PenFor}. In all cases, the penetration depth and the force increased drastically with the increase of the end-effector's velocity, which confirms that the velocity should be constrained to avoid severe harm to the users. The high stiffness case (case (2)) always has the highest penetration depth and force no matter at low or high velocities, which means that rigid actuators would impose higher risks of injury if utilized at similar velocities compared with VSA (cases (1) and (3)). The comparison between case (1) and case (3) also shows that operating at low stiffness prior to the collision would ensure safety better than swiftly switching to low stiffness once the collision is detected; this can be referred to relatively short collision duration time (less than 60 ms) compared with the time to change the stiffness from high to low level ($450 ms$). In Case 1, at a velocity of 0.48 m/s, zero penetration depth and the smallest penetration force have been recorded. Therefore, it would be selected as the maximum operating velocity during the compensation process. Moreover, the stiffness control strategy should ensure that the level of stiffness should be low during most of the SRL trajectory-tracking.

\vspace{-0.2cm}

\subsection{Experimental tracking performance evaluation}

\begin{figure}[t!]
\centering
\includegraphics[width=3in]{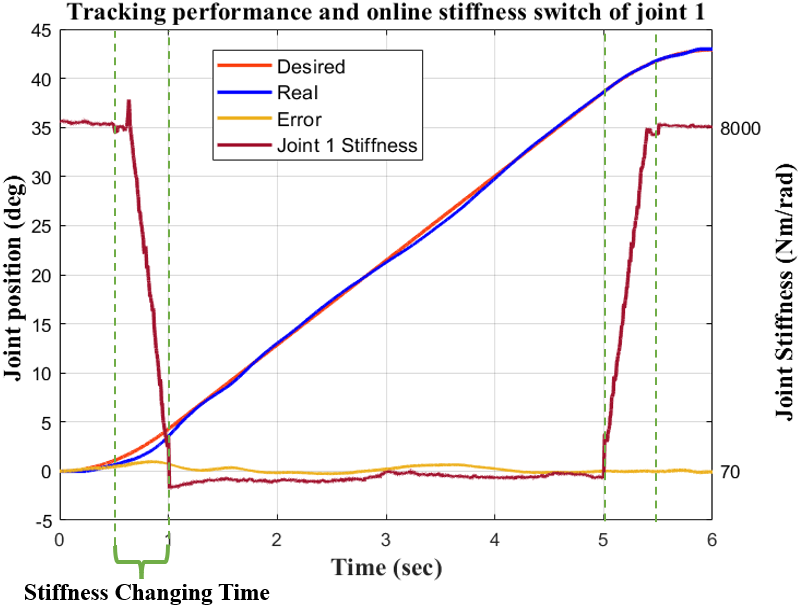}
\caption{Position tracking of joint 1 with a trapezoidal velocity profile (left y-axis), and online stiffness switch of joint 1 (right y-axis).}
\label{fig_ctr1}
\vspace{-0.2cm}
\end{figure}

\begin{figure}[t!]
\centering
\includegraphics[width=3in]{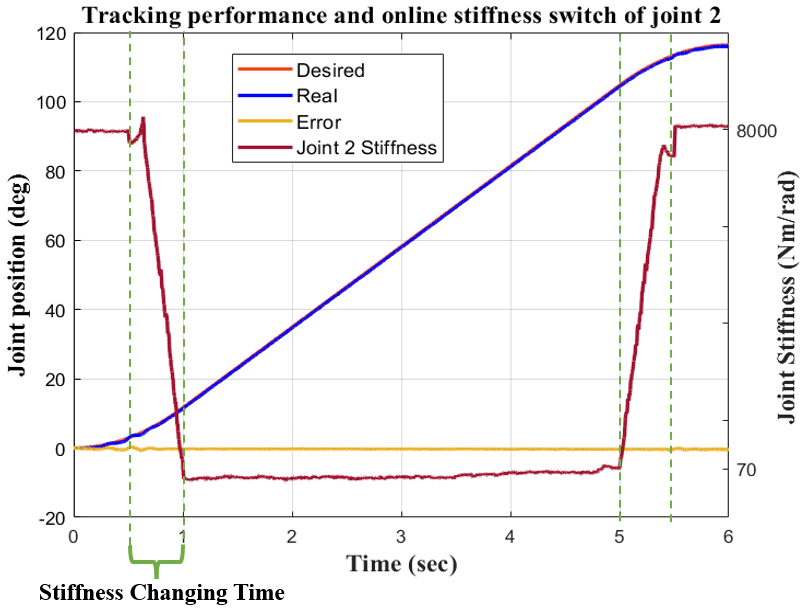}
\caption{Position tracking of joint 2 with a trapezoidal velocity profile (left y-axis), and online stiffness switch of joint 2 (right y-axis).}
\label{fig_ctr2}
\vspace{-0.6cm}
\end{figure}

In this section, the tracking performance of the SRL based on a $PID$ controller is evaluated. The desired final Cartesian position $[x_{df} = -23.62 $ $ mm; y_{df} = 650.69 $ $ mm]$ of the end-effector defined in the base frame is used to calculate the desired final SRL joints $[\theta_{1df};\theta_{2df}]$. A joint trajectory generator based on the trapezoidal velocity profile is adopted to generate the desired trajectory between the initial joint positions and the desired ones. The generated trajectories for both joints, starting from the initial to the final positions, lie within the cooperative workspace defined in Fig. \ref{fig_work} to ensure that the robot links and end-effector do not collide with the user. The trapezoidal velocity profile contains three phases: the acceleration phase where: the acceleration is constant, the velocity is a linear function of time, and the position is a parabolic function of time; the coasting phase where: the velocity is constant and the position is a linear function of time; and the deceleration phase (the negated behavior of the acceleration). This trapezoidal velocity profile allows to control the velocity and imposes a maximum acceleration.
Based on the results of the previous section, the Cartesian end-effector velocity was limited to 0.4 m/s to ensure zero knife penetration in the event of any accidental collision (stabbing) with humans. The maximum joint velocities used for the trapezoidal trajectory generation respect this safety criterion, and their corresponding Cartesian velocities were calculated based on the Jacobian matrix. Moreover, the optimal stiffness profile combined with the trapezoidal velocity profile was adopted as in \cite{Bicchi2004}. Two stiffness levels were used, where the high stiffness is switched at the first half of the acceleration phase (from $0$ to $0.5$ $sec$), and the second half of the deceleration phase (from $5.5$ to $6$ $sec$). The low level of stiffness is switched from the second half of the acceleration phase to the end of the first half of the deceleration period (from $0.5$ $sec$ to $5.5$ sec). The stiffness profile of both joints are illustrated in Fig. \ref{fig_ctr1} and Fig. \ref{fig_ctr2}, the high level of stiffness is set to $8000$ $Nm/rad$, and the low level of stiffness is set to $70$ $Nm/rad$.

 Figures \ref{fig_ctr1} and \ref{fig_ctr2} illustrate the joint positions tracking, where the $rms$ error of joints 1 and 2 are $0.3490$ $deg^2$ and $0.3023$ $deg^2$ and the maximum absolute errors are $0.9835$ $deg$ and $0.6940$ $deg$ respectively. 
 Although a small mechanical backlash on joint 1 slightly degraded the tracking performance, which is affecting the final Cartesian position of the end-effector (at $t=6$ $sec$), the final Cartesian error of the end-effector is less than 3 mm, which is very acceptable for the targeted application. 
\vspace{-0.15cm}

\section{Assistance in Bimanual Tasks}
\label{sec5:qualititative}

In order to evaluate the quality of the assistance in bimanual tasks performed by the proposed SRL, a qualitative questionnaire was conducted. However, prior to the evaluation, a finite state machine was adopted to perform the sequence of assisting the human in eating with a fork and knife. The details are illustrated in the following subsections: 
\vspace{-0.3cm}

\begin{figure}[t!]
\centering
\includegraphics[width=3.5in]{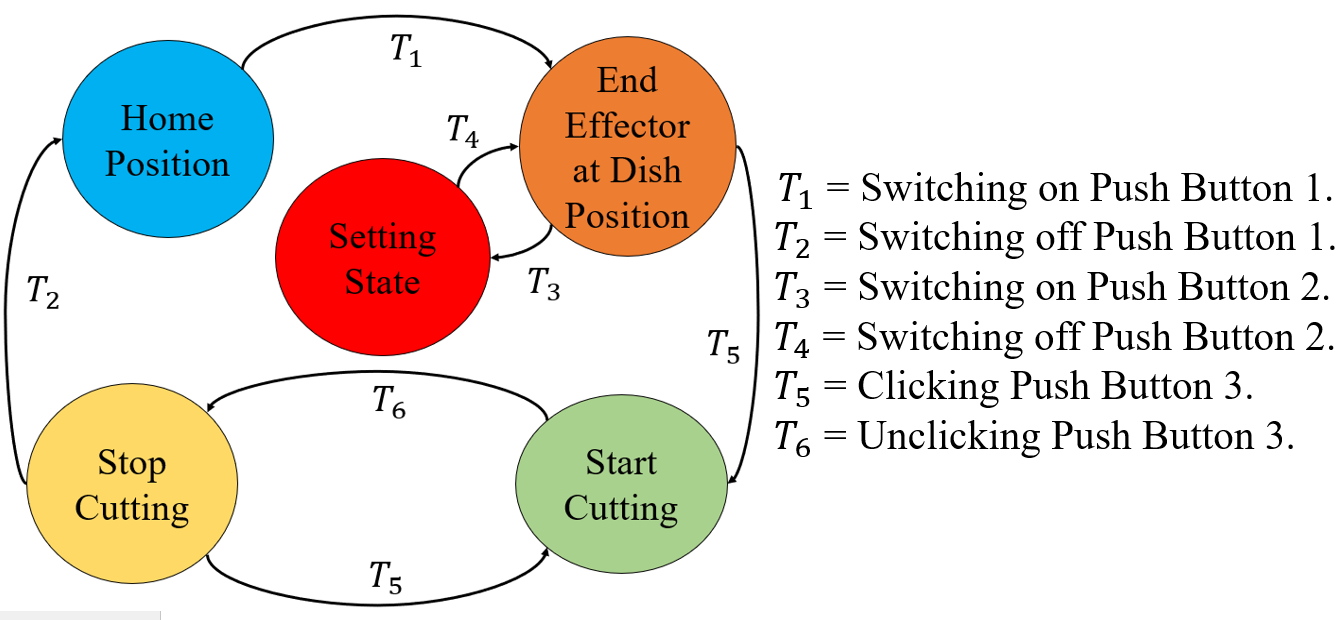}
\hfil
\caption{The FSM describing the operation of the SRL during bimanual eating task.}
\label{fig_sim}
\vspace{-0.5cm}
\end{figure}

\vspace{-0.1cm}

\subsection{Task Description using Finite State Machine (FSM)}

The Finite State Machine (FSM) diagram, depicted in Fig. \ref{fig_sim}, illustrates the control of the collaborative bimanual eating task. 
Initially, the SRL is at the Home Position, representing the first state (S1). At this state, the stiffness of the actuators is set high as long as the SRL is at rest. When the user clicks a push button (B1), the SRL starts moving towards the predefined position of the dish's center (S2). The operational stiffness will be low to ensure safety through the motion between the two states. If the SRL has successfully reached the desired position, the SRL will be maintained at rest, and the stiffness will be set high to ensure accuracy. If the SRL has not reached the desired position, the patient can press the switch (B2), which activates the Setting State (S3). In this state, the SRL would operate in zero-torque mode to allow the user to change the final position to the desired location, which would be recorded. Once the desired position is set, the user switches (B2) off to return to (S2). It is worth mentioning that the user can only alter the end-effector's position within the defined cooperative workspace. In order to initiate the reciprocating knife cutting process (S4), the user would use their foot to activate the push button (B3), which would maintain the cutting process until the user releases the button. The user may send the SRL back to the homing state by switching the homing button (B1) off.

\vspace{-0.1cm}
\subsection{Questionnaire Results}
A subjective qualitative questionnaire according to  \cite{8122720} was conducted on ten healthy subjects (five males and five females, with an age of ($29 \pm 3$) years old to evaluate the usability of the system and the user's overall satisfaction. The purpose of conducting the test on healthy participants is to get initial insights regarding the performance of the proposed device. However, it is intended to be applied to the actual users (hemiplegic patients) for future work. The subjects were given enough time to train on the process before conducting the evaluation. 

The participants were requested to imagine they had only one functional arm, and they had to use the device to compensate for the other arm while eating a meal. In addition, they were requested to declare their level of agreement with the statements in Table \ref{tab:sur} which mainly addressed the following aspects: Perceived Usefulness (PU), Perceived Ease of Use (PEU), Emotions (E), Attitudes (A) and Comfort (C) on a 7-point Likert scale. The system was considered relatively helpful and easy to use based on the collected responses. However, the participants recommended that the system be more intuitive and with the ability of user intention prediction. Therefore, the authors would work on enhancing the control technique by including other sensory systems (vision sensors, and Inertial Measurement Units (IMUs)) to utilize an advanced control strategy.

\begin{table}[t!]
\caption{The table illustrates the results of completing the questionnaire using a 7.0 Likert scale. 
}

\centering
\renewcommand{\arraystretch}{1}
\begin{tabular}{|p{0.15\linewidth} | p{0.6\linewidth} | p{0.1\linewidth}|}
\hline
Perceived outcome & Eating Bimanual Task & Mean (SD)\\
\hline
PU1 & The robot is useful in enhancing task efficiency & 5.8 (0.29)\\
\hline
PU2 & The control strategy of the SRL facilitated the arduous bimanual eating activity & 5 (0.5)\\
\hline
PU3 & The device is precise in doing the prescribed movements & 6.2 (0.58)\\
\hline
PEU1 & It is effortless and straightforward to use the SRL 
& 6.3 (0.29)\\
\hline
PEU2 & It is time-efficient to use the proposed system. & 4.8 (0.58)\\
\hline
E1 & I like the design of the arm. It is appealing to my eyes & 5.5 (0.6)\\
\hline
E2 & I like to receive assistance from the robotic arm.& 6.2 (0.32)\\
\hline
A1 & I think this apparatus can be exploited to accomplish other bimanual tasks 
& 6.5 (0.01)\\
\hline
C1  & Both the design and the control of the extra arm ensure safety, which I can perceive. & 6.0 (0.87)\\
\hline
C2  & Using the device is not restricting my body motions & 6.8 (0.64)\\
\hline
C3  & It feels natural to use the robotic device & 5.0 (0.96)\\
\hline
\end{tabular}
\label{tab:sur}
\vspace{-0.5cm}

\end{table}

\vspace{-0.2cm}
\section{Conclusion and Future Work}

In this paper, the novel supernumerary robotic limb was proposed as a proof-of-concept to assist hemiplegic patients in their daily bimanual activities (i.e., eating with a fork and knife). To the author's knowledge, the proposed SRL is the first supernumerary arm based on variable stiffness actuators. The motivation behind the proposed design is to harvest the advantageous features of rigid and soft actuators in ensuring accuracy and safety, respectively. These features can be realized due to the capability of the VSA to operate at high stiffness to ensure accuracy and low stiffness to ensure safety. The proposed SRL consists of a two-link SRL arm actuated by variable stiffness actuators. The end-effector consists of a passive three-DoFs spherical joint with a knife mounted on a linear actuator. The lengths of the links of the SRL were defined to cover the collaborative workspace. The soft-tissue-injury test was conducted to evaluate the system's safety, and the results revealed that the proposed system would perform safely at higher velocities if compared with rigid-actuator-based SRLs; this is due to the shock absorption feature in the variable stiffness actuators. To evaluate the system's accuracy, trajectory following tests were conducted, and the results revealed high accuracy in trajectory following with minimal tracking error. In order to prepare the system to assist hemiplegic patients in eating with a fork and knife, a finite-state machine was implemented to control the SRL assistance process. The quality of the process was evaluated and proven in terms of usefulness through a subjective qualitative questionnaire conducted on healthy subjects.

The drawbacks of the proposed system lie in the relative bulkiness of the VSAs; the main contributor to the bulkiness is the utilization of relatively large motors with gear heads. However, this drawback can be overcome by realizing more compact designs of VSA and utilizing relatively smaller motors with similar torque capacities. In addition, the authors will explore other control algorithms in the future, including iterative learning control and phase-variable control strategies. Moreover, qualitative experiments with hemiplegic patients will be performed.
\vspace{-0.15cm}
\section*{Acknowledgment}
This work was supported by Khalifa University of Science and Technology under Award RC1-2018-KUCARS. 
\vspace{-0.2cm}
\bibliographystyle{IEEEtran}
\bibliography{root}

\vfill


\end{document}